%% file: FG2015_archive.tex
\documentclass[a4paper, 10pt, conference]{ieeeconf}     
 \pdfoutput=1                                                         
\usepackage{FG2015}                                              
\usepackage{graphicx}
\usepackage{amsmath,amssymb} 
\usepackage{color}
\usepackage{multirow}
\usepackage{array}
\FGfinalcopy 

\title{\LARGE \bf
The Beauty of Capturing Faces: Rating the Quality of Digital Portraits}
\author{\parbox{16cm}{\centering
    {\large Miriam Redi$^1$,  Nikhil Rasiwasia$^2$,  Gaurav Aggarwal$^2$,  Alejandro Jaimes$^3$}\\
    {\normalsize
    $^1$ Yahoo Labs, Barcelona, Spain $^2$ Yahoo Labs, Bangalore, India  $^3$ Yahoo Labs, New York, USA}}
}

\begin{document}

\ifFGfinal
\thispagestyle{empty}
\pagestyle{empty}
\else
\author{Anonymous FG 2015 submission\\-- DO NOT DISTRIBUTE --\\}
\pagestyle{plain}
\fi
\maketitle

\begin{abstract}
Digital portrait photographs are everywhere, and while the number of face pictures keeps growing, not much work has been done to on automatic portrait beauty assessment. 
In this paper, we design a specific framework to automatically evaluate the beauty of digital portraits. To this end, we procure a large dataset of face images annotated not only with aesthetic scores but also with information about the traits of the subject portrayed. We design a set of visual features based on portrait photography literature, and extensively analyze their relation with portrait beauty, exposing interesting findings about what makes a portrait beautiful. We find that the beauty of a portrait is linked to its artistic value, and independent from age, race and gender of the subject. We also show that a classifier trained with our features to separate beautiful portraits from non-beautiful portraits outperforms generic aesthetic classifiers.
\end{abstract}

\input{sec/1_introduction}
\input{sec/2_relatedwork}
\input{sec/3_dataset}
\input{sec/4_features}
\input{sec/5_analysis}
\input{sec/6_results}
\input{sec/7_conclusions}

\bibliographystyle{splncs}
\bibliography{sigproc}

\end{document}

%% file: sec/1_introduction.tex
\section{Introduction}
Portraits make up a large percentage of the photos on the web nowadays.  ``Selfies'' have become a phenomenon, and recent studies~\cite{facesinstagram} show that images with faces are more popular (+38\% ``likes" on Instagram) than other pictures in online social networks. Portraits are also used in web user profiles, in news articles, to represent celebrities and public figures, and they are an essential part of all kinds of IDs. 
%

%
Given the huge volume of digital portraits, their broad usage, and their importance for people identification, surfacing the best digital portraits in terms of  photographic quality is of crucial importance. A system able to automatically score the aesthetic value of portraits could be used to select good images for a variety of applications such as  journalism, photo sharing websites, web search, PhotoBoosts, and many others.

Shooting photos of people is not a trivial task: human faces convey emotions, stories, lifestyles, and a good photographer needs to be able to capture their essence and personality. As a matter of fact, portrait photography is a stand-alone branch of photography literature, with  its own rules and compositional techniques, and tons of dedicated books  \cite{child2008studio, hurter2007portrait, weiser1999phototherapy}. Systems that automatically rate the quality of digital portraits should be therefore specifically designed for face photos, unlike traditional visual aesthetics works \cite{datta, ke2006design}, based on general photographic rules .

In spite of its importance, there has been little work in the research community to specifically address computational aesthetics of portraits.  Preliminary works ~\cite{khan2012evaluating, li2010aesthetic} leave out many of the aspects that are specific to portraits (e.g., illumination, landmark representation, affective properties, etc.), and have experimented only with small datasets (less than 500 images). 

\begin{figure}[t]
\centering
\includegraphics[width=\linewidth, natwidth=1362, natheight=921]{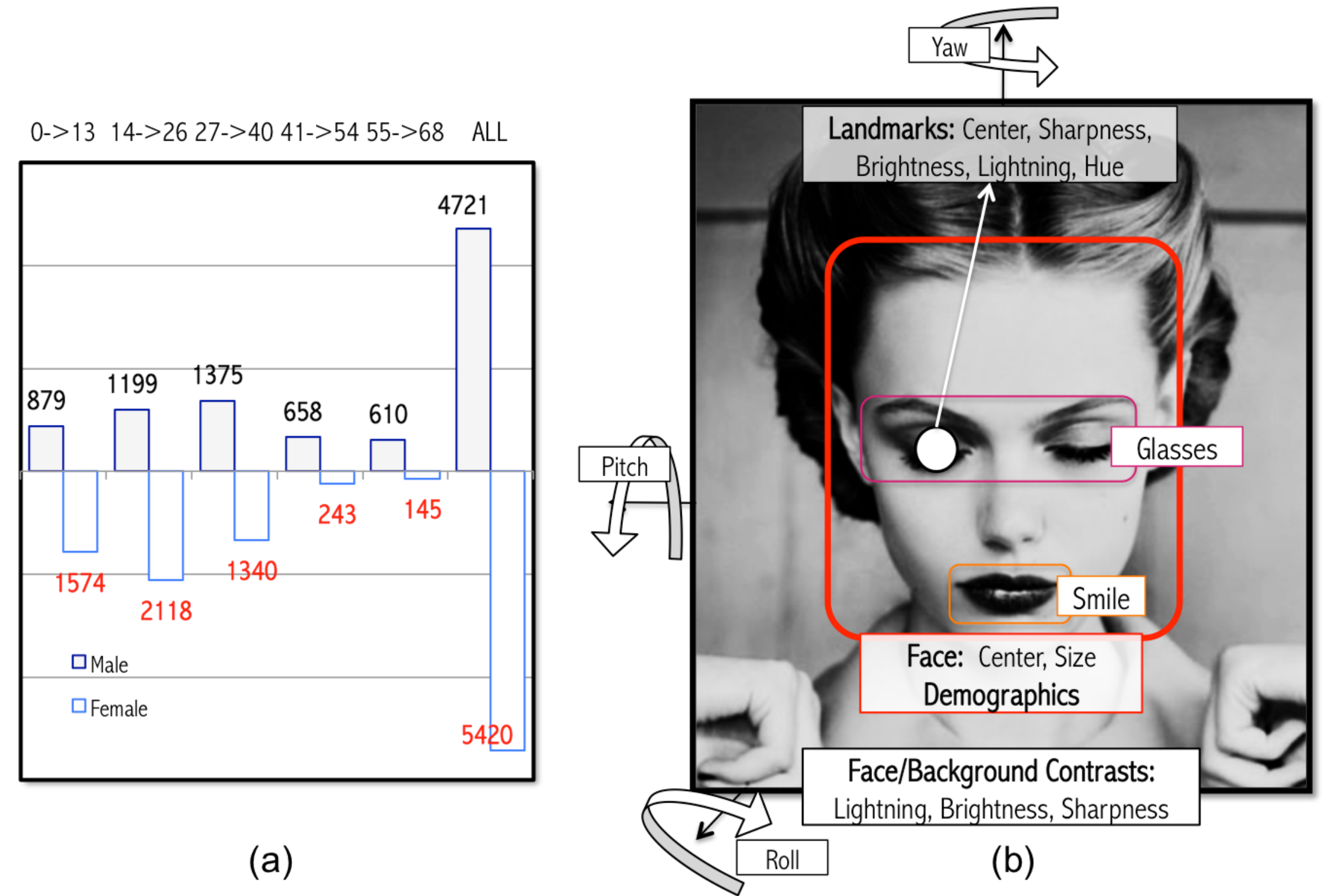}
\caption{(a) Distribution of images per demographic category and aesthetic scores (b) Characteristics extracted from the portrayed subjects}
\label{Facefeatures}
\vspace{-0.5cm}
\end{figure}
In this paper, we try to fill this void and introduce a new framework to automatically evaluate portrait aesthetics\footnote{The aim of this work is to estimate the photographic quality of the \textit{representation} of the person, independent from the beauty of the subject represented. }. To do so, we design visual features to describe image quality and portrait-specific properties and present a large-scale analysis of a data set of  over 10,000 portraits.
In addition, we build predictive models that are able to determine the aesthetic score of digital portraits.
Moreover, with such large scale study, we provide an analysis of what makes a portrait beautiful from a computational perspective. To our knowledge, this represents the first attempt in literature to understand the relevancy of features for portrait aesthetics. 

Our main contributions can be summarized as follows:
\\\textbf{(1) Dataset:} we build a large dataset of portraits annotated with physical characteristics (determined using facial analysis) by sampling the AVA~\cite{murray2012ava} images.
\\\textbf{(2) Features:} we introduce new features to describe portrait composition, quality, illumination, memorability, emotions, and originality.
\\\textbf{(3) Feature analysis:} we perform analyses on a set of over 10,000 portraits and report observations. We find that race, gender, and age are largely uncorrelated with photographic beauty, but aesthetic score is related to sharpness of facial landmarks, image contrast, exposure, homogeneity, illumination pattern, uniqueness, and originality.
\\\textbf{(4) Aesethetic Prediction:} we develop predictive models to classify portraits as aesthetically beautiful or not.

In Sec. \ref{related}, we describe related work, and explain our portrait dataset in Sec. \ref{datasets}. Sec. \ref{visual_features} presents the visual features and analyze their relations with portrait beauty in Sec. \ref{analysis}, then present our classification experiments in Sec. \ref{results}. %

%% file: sec/2_relatedwork.tex
\section{Related Work}\label{related}
Our work relates to research that applies image analysis techniques to  detect the visual presence of non-semantic, fuzzy concepts such as memorability~\cite{isola2011}, emotions~\cite{emotions,joshi2011aesthetics}, interestingness~\cite{redi2012interestingness, interesting, gygli2013interestingness}, privacy~\cite{zerr2012privacy}, and beauty~\cite{datta}.
 In particular, this paper follows previous work on computational aesthetics~\cite{datta,murray2012ava}, that explores the discriminative ability of visual features to automatically assess the beauty of images and videos. Pioneers in this field are~Datta et al. and Ke et al., \cite{datta, ke2006design}, who built an aesthetic classification framework for images based on features inspired by photographic theory. In subsequent years, such works were improved by designing more discriminative features~\cite{nishiyama2011aesthetic, interesting}, proving the effectiveness of generic features~\cite{marchesotti2011assessing, murray2012ava}  and building more effective learning frameworks~\cite{wu2011learning}. Similar frameworks were applied to automatic image composition and enhancement by Bhattacharya et al.~\cite{bhattacharya2010framework}.

While these existing computational aesthetic works build general frameworks for photographs of any semantic category, we focus on a specific type of images, namely portraits, whose compositional  and aesthetic criteria constitute a separate subject of study in the photographic literature~\cite{hurter2007portrait, weiser1999phototherapy, child2008studio}, and therefore need a separate computational framework for aesthetic assessment.
This aspect is also proven by our experiments: we show that our portrait-specific aesthetic framework performs much better than a general classifier  for portrait aesthetic assessment.

A few works~\cite{murray2012ava, luo2011content, interesting} perform topic-based aesthetic classification. They build category-specific subsets of images by sampling aesthetic databases according to given image tags (``city", ``nature", but also  ``humans" or ``portraits"), and then use general compositional features to build topic-specific models.  The framework in this paper differs from those works for two reasons. (1) \textit{We build a rich large-scale portrait aesthetic database}. A dataset based on tag-based sampling as in \cite{murray2012ava, luo2011content, interesting}, could ignore many face images without tags while including images with noisy tags (as shown in Section \ref{datasets}).  In this paper, we adopt a \textit{content-aware sampling strategy} based on detailed face analysis. We reduce a large scale aesthetic dataset \cite{murray2012ava} to a subset of more than $10000$ face images annotated with information about the portrayed subject, useful for both analysis and feature extraction. (2) \textit{We build portrait-specific aesthetic visual features}. The works in~\cite{murray2012ava, luo2011content, interesting} use traditional aesthetic  features designed for a general case, and apply them to the topic-specific contexts. In our work, we design face-specific aesthetic features inspired by photographic literature, together with non-face features that describe crucial aspects of photographic portraiture, such as illumination, sharpness, manipulation detection, image quality, emotion and memorability. Moreover, we show their combined effectiveness for aesthetic assessment of face photographs compared to  traditional aesthetic features. 

There are a few recent works that attempt to design portrait specific datasets and features. For example, Li et al.\cite{li2010aesthetic} use  face expression, face pose and face position features to estimate the aesthetic value of the images in a dataset of $500$ face images annotated by micro-workers. This work was improved by the work in~\cite{xuefeature}, that uses hand-crafted features together with low-level generic features, and by Khan et al.~\cite{khan2012evaluating} using spatial composition rules specifically tailored for portrait photography, together with specific background contrast features and face brightness and size features. These works represent a first attempt towards portrait aesthetic classification. However, one major weak point of such works is that they rely on small datasets (<500 images), thus making the results less generalizable for large datasets like the one we consider. Moreover,  despite their focus on face analysis, the features proposed by those works miss many important aspects of portrait photography such as illumination, demographics, face landmark properties, affective dimension, semantics and post-processing. In our work, we use features that are able to capture these aspects, and prove their effectiveness  by showing that they outperform the features in \cite{khan2012evaluating} when used in an aesthetic classification framework on the dataset used by Khan et al.~\cite{khan2012evaluating}. Moreover, in this paper, we perform for the first time a deep analysis of the importance of each feature and each group of features for face photo aesthetics, giving interesting and probably unexpected insights about what makes a portrait beautiful.
%

%% file: sec/3_dataset.tex
\section{Large Scale Portrait Dataset}\label{datasets}
In order to create a large scale corpus of face images annotated with beauty scores, we resort to the largest aesthetic database available in the literature, i.e. the AVA dataset \cite{murray2012ava}, created from the photo challenge website dpchallenge.com, that contains more than $250,000$ images annotated with an aesthetic score, a challenge title, and semantic textual tags.

AVA is a unique, rich dataset for visual aesthetics, and therefore a reliable source of data for our purposes.
However, AVA images contain very diverse subjects other than faces.  Moreover, for analysis and classification purposes, we want to collect not only a reliable subset of portrait images, but also some rich information about the portrayed subject and its representation. With this in mind, we design a \textit{content-aware sampling strategy} on the AVA dataset, based on both metadata-based filtering and face analysis:
\\(1) \textbf{Enhanched metadata-based filtering.} First, we select  from the AVA database not only the images tagged as ``Portrait'' but  also  all the images whose challenge title contains the words 'Portrait', 'Portraiture' or 'Portraits'. (e.g. \textit{Portrait Of The Elderly}). A total of 21,719 images are collected at this stage.
\\(2) \textbf{Face detection-based filtering.}  We use  Face++
~\cite{face++} to filter the images collected after metadata-based filtering. We obtain  a subset of 10,141 images for which Face++  detected the presence of one or more faces (in case of multiple faces, we retain the information about the largest one only).
\\(3) \textbf{Subject properties.} We compute though Face++ basic information about the subject, such as position, orientation, demographics (race, gender, age), coordinates of facial landmarks (eyes, nose and mouth in relative coordinates), presence of smile, presence of glasses, etc. (for a complete list of features see Table \ref{tab:features}). 

For each of the resulting images, we assign the average aesthetic score (in a 1-10 range) according to the votes provided by the AVA dataset. Figure \ref{Facefeatures} shows the composition of our dataset, highlighting the distribution, based on gender and other properties estimated by the Face++ detector. About 53\% of the subjects are classified as female, and 1/3 of the image corpus shows subjects between 14 and 26 years of age (Fig. \ref{Facefeatures} (a)). Similar to the AVA dataset, the vast majority of the aesthetic scores lies between 4 and 6, with a peak around the mean, which stands at 5.5.

%% file: sec/4_features.tex
\section{Features for Portrait Aesthetic Assessment}\label{visual_features}
\input{tables/features}
Visually stunning portrait photographs are often  the result of an artistic process that might not strictly follow  general rules of composition, or fulfill basic quality 	requirements. However, photographic portraiture literature
 \cite{child2008studio, hurter2007portrait, weiser1999phototherapy} suggests that following some specific photographic principles can help making digital portraits more attractive, ensuring visual appeal and expressiveness. Among the various tips for good portraiture available in literature, we identified 5 main photographic dimensions, namely:
\\\textbf{Compositional Rules:} arrangement of lines, objects, lights and color, widely used in visual aesthetic literature \cite{datta, luo2011content}.
\\\textbf{Scene Semantics:} where has the photo been shot? and which objects co-exist with the subject in the scene?
\\\textbf{Portrait-Specific Features:}  information about the subject (aspect, soft biometrics, demographics) and its representation (sharpness, illumination, etc.)
\\\textbf{Basic Quality Metrics:} principles that ensure the correct perception of the signal, without distorting the scene represented.  Rarely used in computational aesthetics, they can be fundamental for high-quality portraiture \cite{hurter2007portrait}.
\\\textbf{Fuzzy Properties:}  portrait photographic beauty is related to non-objective properties such as emotions or uniqueness, which are unquantifiable with low level features. 

In this work, we design 5 groups of features that aim at describing various aspects of each of these dimensions using computer vision techniques. 
\subsection{Compositional Rules} \label{sec:comp}
As highlighted in many previous works \cite{datta, pere, bhattacharya2010framework}, the visual attractiveness of a  picture is strongly influenced by the arrangement of objects in the image, their lighting, their colors,  their perceptibility.

Similar compositional rules apply to portraits photography. However, since portraits generally focus on a single subject whose essence needs to be captured in the shot, two compositional aspects need particular consideration: lighting and sharpness. The correct illumination of the scene and the detailed representation of the subject ensures both  perceptibility and expressiveness. Given these observations, we design a set of new features that capture essential properties of image lighting and sharpness, and collect a set of existing features for image composition analysis.
\vspace{0.5cm}
\\\emph{Lighting  Features} \\
The lighting setup is crucial to determine the essence of the portrait. In previous works \cite{datta,khan2012evaluating, luo2011content}, scene lighting is described using features based on overall image brightness. However, as proved by our results, the raw brightness channel information might not be enough to capture portrait lighting patterns. 

We therefore design a new lighting feature to expose \textbf{Lighting Patterns} based on  an illumination compensation algorithm originally created for face recognition \cite{gross2003image}.  Such method considers an image $I$  as a product $I=R(I)\cdot L(I)$, where $R(I)$ is the  'reflectance' of the image and $L(I)$ is its  ``illuminance'" i.e. the perceived lighting distribution.

\begin{figure}[t]
\centering
\includegraphics[width=\linewidth,natwidth=1745,natheight=336]{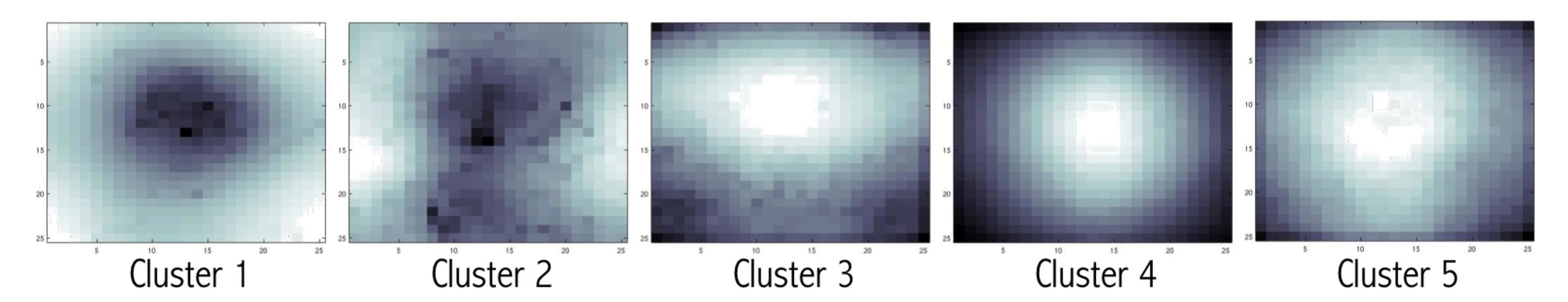}
\caption{Illuminance Distribution of the 5 Lightning Patterns}
\label{fig:otherfea}
\vspace{-0.5cm}
\end{figure}

In order to infer the lighting  pattern of an image, we proceed as follows.
For each image, we calculate $L(I)$ and create an illuminance vector $V(I)$ by averaging its  illuminance $L(I)$ over local windows (25x25 subdivision).  Applying k-means clustering on the illuminance vectors of a set of training images, we group the illuminance vectors into 5 Lighting Patterns representing the most common lighting setups in our dataset (See Fig.\ref{fig:otherfea}).
For a new image $J$, we assign its corresponding lighting pattern by looking at the closest cluster to its illuminance vector $V(J)$, and retain the cluster number as the \textit{Lighting Pattern Feature}.
\vspace{0.2cm}
\\\emph{ Sharpness  Features} \\
The recognizability and sharpness of the subject is a basic requirement for good portraiture. To analyze the amount of sharpness in the image, we design two new features:
\\\textbf{Overall Sharpness}: 
Subject movements or camera defocus can affect the overall image sharpness, introducing disturbing blur in particular image regions. We compute the sharpness of a picture by calculating the strength of the edges after applying horizontal and vertical Sobel masks on the image, according to the Tenengrad method (as explained in \cite{ng2001practical}).
\\\textbf{Camera Shake}: sometimes camera movements can create an overall blurriness in the image. In order to estimate this particular type of blur, we compute  the ratio between the number of pixels detected to be affected by camera shake and the total number of pixels, according to the camera motion estimation algorithm of Chakrabarti et al. \cite{chakrabarti2010analyzing}.
\vspace{0.2cm}
\\\emph{Traditional Compoisitional Features}\\
We collect here a set of features from state-of-the art works that model compositional photographic rules using a computational approach.
\\\textbf{Color Features.} In order to capture color patterns and their relation with portrait aesthetics, we compute the following features extracted from literature: \textit{Color names} \cite{emotions}, \textit{Hue, Saturation, Brightness (HSV)} \cite{emotions, datta} , the \textit{Pleasure, Arousal, Dominance} metrics \cite{emotions}, the \textit{Itten Color Histograms} \cite{emotions}, and the corresponding \textit{Itten Color Contrasts}: \cite{emotions} Moreover, we compute 2 contrast metrics: 
\textit{Contrast (Michelson)} \cite{Michel}, and a traditional \textit{Contrast} measure computed as the ratio between the difference of max-min values of the Y channel and the Y average.
\\\textbf{Spatial Arrangement Features.} The distribution of textures, lines and object in the image space is an important cue for aesthetic and affective image analysis, as proved in \cite{datta, emotions, pere, interesting}. 
To analyze spatial layout of objects and shapes in the scene, we  compute first two symmetry descriptors, namely \textit{ Symmetry (Edges)}\cite{interesting}, and \textit{Symmetry (HOG)}, for which we retain the difference between the HOG \cite{dalal2005histograms} descriptors from left half of the image, and  from the flipped right half. Moreover, we compute 2 new features that describe shapes and their distribution, namely the \textit{Number of Circles}, and the \textit{Rule of Thirds}, that, unlike previous works \cite{datta, bhattacharya2010framework}, determines the rule of thirds by computing the amount of spectral saliency \cite{hou2007saliency} in the 9 quadrants resulting from a 3x3 division of the image. 
\\\textbf{Texture Features.} Textural features can help analyzing the overall smoothness, order and entropy of the image. We analyze image homogeneity by computing the \textit{GLCM properties} \cite{emotions}, the \textit{Image Order} \cite{redi2012interestingness}, and the \textit{Level of Detail} \cite{emotions}.
\subsection{ Semantics and Scene Content}
As proved by various works in visual aesthetics \cite{redi2012interestingness, luo2011content, pere}, the content of the scene and the types of objects placed in the picture substantially influence the aesthetic assessment of pictures. In particular, in the portraiture context, it is important to analyze the setting where the photo has been shot, i.e. objects, scenery and overall harmony of subject with the scene. In order to estimate these properties, we compute an adapted version of the \textbf{Object bank features} \cite{li2010object} that retains the maximum probability of a pixel in the image to be part of one of the 208 objects in the Object Bank.
\subsection{Basic Quality Metrics}
In general, visually appealing portraits are also high-qiuality photographs, i.e. images where the degradation due to image registration or post-processing is not highly perceivable.
In order to deeply analyze this dimension, we design some rules to determine the perceived image degradation by looking at simple image metrics, independent of the composition, the content, or its artistic value, namely:
\\\textbf{Noise}: we compute the amount of camera noise by applying an image denoising algorithm \cite{buades2005non}, and then computing the distance between the denoised image and the original one. 
\\\textbf{Contrast Quality}: well-contrasted images, i.e. images where the contrast level allows to distinguish the picture shapes without  introducing disturbing over-saturated regions, can be recognized by the uniform distribution of the intensities on the image histogram.  We therefore compute the quality of the contrast  by negating \footnote{We take the negative of the distance in order to have higher values of this features for higher contrast quality} of the distance between the original image and its contrast-equalized version. 
\\\textbf{Exposure Quality}: the luminance histogram of an overexposed image is skewed towards the right part, while for an underexposed image it is skewed towards the left side. In order to capture this behavior, we convert the image to the YCbCr space,  we compute the skewness of the Y channel histogram over 255 bins. When the skewness is close to zero, the exposure is correct, when below or above zero, the image is under or over exposed. We negate the absolute value of the skewness as exposure balance metric.
\\\textbf{JPEG Quality}: when too strong, JPEG compression can cause disturbing effects such as blockiness or block smoothness. We implement the objective quality measure for JPEG images proposed by \cite{wang2002no} and retain the JPEG quality score output by the algorithm.
\\\textbf{Image Manipulations}: more and more, digital pictures are post-processed after the shooting using editing tools. In order to understand the amount of post-processing applied on the image, we design 2 new quality metrics, inspired by blind image forensics techniques. First, we design a feature to compute the amount of  \textit{Splicing Manipulation}: we retain the output of an SVM classifier trained with Markov Features \cite{pevny2010steganalysis} computed on a training set of images annotated as spliced/not spliced from the CASIA dataset \cite{casia}  (85\% accuracy on this set). 
Next, we build a feature to compute the amount of \textit{Median Filtering Manipulation}, using the algorithm of Yuan et al. \cite{yuan2011blind}.
\subsection{Portrait-Specific Features}\label{facefeatures}
In photographic portraiture, lot of effort should be spent on understanding the subject and its correct representation. Photographic portrait theory \cite{hurter2007portrait} particularly stresses the importance of the focus, sharpness, lighting and position of the face landmarks (eyes,nose,mouth). 

In order to describe the properties of the subject and its representation, we retain as candidate features all the values extracted automatically by the Face++ api, and we build on top of such values a set of features to deeply describe the face and landmark properties. Overall, the set of Face/Subject features is as follows:
\\\textbf{Face++ description:} \textit{Face Position}, namely x, y relative coordinates, plus relative width and height , \textit{Face Orientation}, i.e.,  yaw, pitch and roll angle of the head, \textit{Demographics  } like Race (white, black, asian), Age (in years) and Gender, \textit{Landmark Coordinates}, namely Right/Left Eye, Nose and Mouth position in relative coordinates, \textit{Subject Expression  }, that  estimates wether the subject is smiling or not, and \textit{Other Face Properties } such as  presence of glasses (none, sunglasses, normal glasses). \\
\textbf{Landmark Sharpness} for each landmark, we simply compute its sharpness by averaging the gradient magnitude over the landmark region.
\\ \textbf{Landmark Staitstics:}  for each landmark, we extract its average Hue and Brightness
\\ \textbf{Face/Background Contrasts:} similar to the background contrast feature in \cite{khan2012evaluating}, we analyze here the compositional differences between face region and background region. However,  while Khan et al.  \cite{khan2012evaluating} simply retain the ratio between face region brightness and image brightness, we perform here a deeper analysis. We consider face ($F$) and background ($B$) as two separate sub-images. We then  compute the \textit{Lighting Contrast} as the ratio between the average Lightning (see Sec. \ref{sec:comp}) of F and the average Lightning of B,  the F/B \textit{Sharpness Contrast} (Sharpness is computed computed as for the Landmark Properties), and, similarly, the \textit{Brightness Contrast}.
\subsection{ Fuzzy Properties}
Some artistic traits of photographs cannot be directly captured by low-level features:
many times, photographic beauty is related to feelings vehiculated by the image, which not even words can describe. In our work, we try to model some of those 'fuzzy' properties using a computational approach, by re-using existing work on image memorability, originality and affective analysis.
\\\textbf{Emotion}, is the emotion aroused by the image positive or negative?  We address this question by training an emotion classifier (SVM, 75\% accuracy) with traditional Compositional Features, using as a groundtruth a mixture of 3 affective dataset \cite{dan2011geneva, borth2013large, emotions}. We binarize the annotation in order to reflect the positive/negative trait of the emotion shown. For each image , we retain the emotion score predicted by such classifier as the image emotion feature.
\\\textbf{Originality} of the image composition is computed by retaining the output of an originality classifier trained with Compositional Features and the Photo.net database from \cite{datta} (Support Vector Regression (SVR), 4,7\% MSE ). 
\\\textbf{Memorability} of the image content. We compute this by retaining the output of a memorability classifier trained with the Saliency Moments Features\cite{sm}, and the  memorability database of Isola et al.  \cite{isola2011} (SVR, 2\% MSE). 
\\\textbf{Uniqueness}: as in \cite{redi2012interestingness}, we estimate the photo uniqueness as the euclidean distance between the average spectrum of the images in a database and the spectrum of each image.

%% file: tables/features.tex
\begin{table*}[t]
\small
\resizebox{\textwidth}{!}{
	\begin{tabular}{|m{4.6cm}|c|m{13cm}|c|}
\hline
\textbf{Feature} & \textbf{Dim} & \textbf{Description} & \textbf{References} \\ \hline
\multicolumn{4}{|c|}{\textit{Compositional Features}}\\ \hline
\textit{Lightning Patterns} &5& Lightning pattern according to the image illuminance & new \\  
\textit{Overall Sharpness} &1& Sum of the image pixels after applying Sobel masks & new \\  
\textit{Camera Shake} &1& Ratio between 'moving' pixels identified by the method in \cite{chakrabarti2010analyzing} and image size & new\\ 
\textit{Color Names} &9& Number of pixels that belong to given color clusters such as \textit{black, blue, green, flesh, magenta, purple} & \cite{emotions}  \\ 
\textit{HSV average}  &6& Average \textit{Hue, Saturation, Brightness} of the \textit{whole} image and in the \textit{inner quadrant}  & \cite{emotions, datta}\\ 
\textit{Pleasure, Arousal, Dominance} &3& Affective dimensions computed by linearly combining HSV values  & \cite{emotions}\\ 
\textit{Itten Color Histograms}&20& Histograms of H, S and V values quantized over 12, 3, and 5 bins & \cite{emotions}\\  
\textit{Itten Color Contrasts}&3& Standard deviation of the Itten Color Histograms distributions &  \cite{emotions}\\ 
\textit{Contrast (Michelson)} &1& Ratio between the sum of max and min luminance values and their difference &  \cite{Michel} \\ 
\textit{Contrast} &1& Ratio between the sum of max and min luminance values and the average luminance & new\\ 
\textit{Symmetry (Edge)}&1& Distance between edge histograms on left and right halves of the image &  \cite{redi2012interestingness} \\
\textit{Symmetry (HOG)} &1& Difference between HOG features on left and right halves of the image & new\\
\textit{Number of Circles}&1&Computed using Hough transform & new\\
\textit{Rule of Thirds}&9&Based on saliency distribution of the 9 image quadrants resulting after a $ 3\times3$ division of the image & new\\
\textit{GLCM Properties}&4& \textit{Entropy, Energy, Homogeneity, Contrast} of the GLCM matrix &  \cite{emotions}  \\
\textit{Image Order} &2& Order values obtained through Kologomorov \textit{Complexity} and Shannon's Entropy & \cite{order, redi2012interestingness}\\
\textit{Level of Detail}&1&Number of regions after Watershed segmentation & \cite{emotions}\\ \hline
\multicolumn{4}{|c|}{\textit{ Semantics}}\\ \hline
\textit{Object Bank Features}&189& Object Bank image representation & \cite{li2010object}  \\ \hline
\multicolumn{4}{|c|}{\textit{Basic Quality Metrics}}\\ \hline
\textit{Noise} &1& Distance between original image and image denoised with the algorithm from \cite{buades2005non}& new \\  
\textit{Contrast Quality} &1& Negative distance between original image and image with normalized contrast& new\\ 
\textit{Exposure Quality} &1& Negative absolute value of the luminance histogram skewness& new\\ 
\textit{JPEG Quality} &1& Computed with the no-reference quality estimation algorithm in  \cite{wang2002no}& \cite{wang2002no}\\ 
\textit{Image Manipulations} &2& Amount of \textit{Splicing} and \textit{Median Filtering} applied to the image & new\\ \hline
\multicolumn{4}{|c|}{\textit{Portrait-Specific Features}} \\ \hline
\textit{Face Position \cite{face++}}&4& \textit{X, Y} in relative coordinates, plus relative \textit{Width} and \textit{Height}  & \cite{face++}
\\ \textit{Face Orientation  \cite{face++}}&3&  \textit{Yaw}, \textit{Pitch} and \textit{Roll} angle of the head & \cite{face++}
\\ \textit{Demographics  \cite{face++}}&6& Race (\textit{White, Black, Asian}), \textit{Age} (in years) and \textit{Gender} & \cite{face++}
\\ \textit{Landmark Coordinates  \cite{face++}} &8&  \textit{Right}/\textit{Left} Eye, \textit{Nose} and \textit{Mouth} position in relative coordinates & \cite{face++}
\\ \textit{Similing Expression  \cite{face++}}&1&  Estimates wether the subject is smiling or not & \cite{face++}
\\ \textit{Other Face Properties  \cite{face++}} &3&  Presence of \textit{Glasses} (none, sunglasses, normal glasses) & \cite{face++}\\
\textit{Landmark Statistics}&12& \textit{Hue} and \textit{Brightness} of Right/Left Eye, Nose and Mouth & new\\
\textit{Landmark Sharpness}&4& Sharpness of  \textit{Right}/\textit{Left} Eye, \textit{Nose} and \textit{Mouth} using gradient magnitude  & new \\
\textit{Face/Background Contrasts:}&3& Contrast between face region and background in terms of \textit{Lightning}, \textit{Sharpness} and \textit{Brightness} & new \\\hline
\multicolumn{4}{|c|}{\textit{ Fuzzy Properties}}\\ \hline
\textit{Emotion}&1&  Estimates the positive/negative traits of the emotions that the image arouses using compositional features and affective image datasets \cite{borth2013large,dan2011geneva,emotions}  & new
\\ \textit{Originality} &1&  Estimates the image originality based on a classifier trained on the Photo.net dataset \cite{datta}  & new\\
\textit{Memorability}&1& Estimates the image memorability based on a classifier trained on the memorability dataset \cite{isola2011}  & new\\
\textit{Uniqueness} \cite{redi2012interestingness} &1& Based on the image spectrum  & \cite{redi2012interestingness} \\ \hline
\end{tabular}}
 \caption{Visual features for portrait aesthetic modeling}
 \label{tab:features}
 \vspace{-1cm}
 \end{table*}

%% file: sec/5_analysis.tex
\section{What Makes  a Portrait Beautiful?}\label{analysis}
Among all the features in Section \ref{visual_features}, which of them is more discriminative to identify beautiful  portraits in a computational framework? In this Section we explore the relations between the visual features extracted and portrait aesthetic scores, by first analyzing the importance of each feature group described in Sec \ref{visual_features}, and by then looking at the relevance of each  single feature within dimensions defined.
\subsection{Feature Groups for Portrait Aesthetics}\label{multicorr}
To measure the significance of the five feature sets, we perform regression analysis using LASSO~\cite{tibshirani1996regression}
for the different groups of features (i.e. \textit{Compositional Features}). Once the regression parameter vector is
learned, we use compute the Spearman correlation between the predicted scores and the original aesthetic scores. This gives us a multidimensional correlation
metric that indicates the relevance of feature group
for portrait aesthetic assessment.  
We split the data into 5 random partitions, using one of the partitions as the test set
and the rest as training, and
learn regression coefficients to predict the aesthetic scores on the test set using the different groups of features.

As shown in Fig. \ref{fig:analysis}(b) all the groups of features  correlate positively with aesthetic scores. As expected, given the importance face of representation for portraiture, the \textit{Portrait-Specific Features} correlate the most among all the  groups of features proposed, with a correlation of $0.330 \pm 0.029$.  
Despite its rich semantic analysis, and the proved effectiveness for scene
analysis \cite{li2010object}, the ObjectBank \textit{Semantic Features}, with its
$~190$ feature detectors, are not as predictive, achieving a correlation score of
$0.211 \pm 0.022$ in contrast to compositional features which achieve a
correlation score of $0.290 \pm 0.029$. In comparison to these large
feature sets, smaller sets of features such as \textit{Basic Quality} and \textit{Fuzzy
Properties} with $6$ and $4$ dimensions respectively achieve a much lower correlation
score for portrait aesthetics assessment as a whole, despite the importance of
single features within the groups.

In order to calculate the combined predictive power of the whole set of features proposed, we perform similar regression analysis on all features together , i.e. without logical grouping, and look at the behavior of the algorithm  as more and more features are taken into account.
Figure \ref{fig:analysis}(d) shows a plot of the Spearman correlation of the feature set as a function of the number of features used and chosen by LASSO.   Using one single feature (\textit{Right\_Eye\_Sharpness}), the Spearman correlation between predicted aesthetic scores and original aesthetic scores is $0.252 \pm 0.018$. The best correlation score of $0.398 \pm 0.027$ is obtained taking into account all $~300$ features. However, adding more than $60$
features shows diminishing returns. The correlation with $60$ features stand at
$0.37$. The smallest mean square error achieved on the test set stands at
$0.430 \pm 0.008$. 

Table~\ref{tab:featuresrank} reports the weight of the features ranked by when they are first picked by LASSO. Also
reported are the feature category and weights. Notice how all the feature groups appear in the top-10 features, thus confirming the importance of each dimension we consider for portrait aesthetic evaluation, with a predominance of face features. We can  also spot some first insights about the importance of single features: crucial for aesthetic prediction are  landmark sharpness (\textit{Right\_Eye} and \textit{Left\_Eye}), the Exposure Quality, and the high discriminative ability of the Fuzzy Properties \textit{Uniqueness}.
%
%
%
%
%
%
\begin{table}
\scriptsize
\centering
\begin{tabular}{|c|c|c|c|}
\hline
\textbf{Rank} & \textbf{Feature Name} & \textbf{Feature Group} &\textbf{Weight} \\\hline
1 & Left\_Eye\_Sharpness & Portrait Features & \bf{0.061894} \\
2 & Right Eye\_Sharpness & Portrait Features & \bf{0.074302} \\
3 & Exposure\_Balance & Basic Quality & -0.031212 \\
4 & Uniqueness & Fuzzy Properties & \bf{0.14232} \\
5 & Smiling & Portrait Features & -0.045702 \\
6 & Cluster4\_Lighnting & Compositional & 0.017803 \\
7 & Fence & Semantics & -0.022525 \\
8 & Hue\_Inner\_Quadrant & Compositional & -0.045009 \\
9 & Nose\_Hue & Portrait Features & -0.03898 \\
10 & Flower & Semantics & 0.026438 \\
\hline
\end{tabular}
\caption{Feature ranks based on Lasso regression}
\label{tab:featuresrank}
\vspace{-1cm}
\end{table}
\subsection{Single Features for Portrait Aesthetics}\label{singlecorr}
To analyze in a more detailed manner which features correlate most with beautiful portraits, we partition the dataset into 5  subsets, as in Sec. \ref{multicorr} and  average the Spearman correlation coefficient $\rho$  between the individual features values and the aesthetic scores of each partition. 

\begin{figure*}[t]
\centering
\includegraphics[width=\linewidth,natwidth=1008,natheight=288 ]{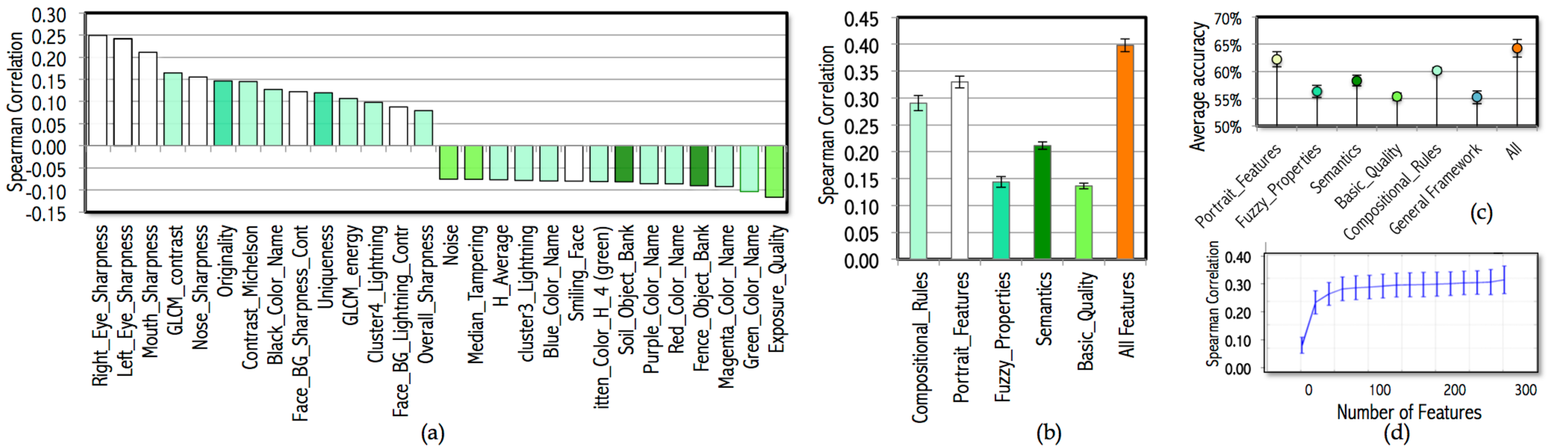}
\caption{Analysis of the most relevant features and components for portrait aesthetic prediction (a,b,d). Classification Performances (c).}
\label{fig:analysis}
\vspace{-0.5cm}
\end{figure*}

In Figure \ref{fig:analysis} (a), we report the $\rho$ coefficients of the features that show higher correlation with portrait aesthetics. We can notice how face sharpness and lighting  are of crucial importance for portrait beauty, as suggested by the Lasso analysis of discriminative features, and by portrait aesthetic literature.  4 out of the top 5 positively-correlated features correspond to the landmark sharpness features. Also, the contrast in sharpness between face and background strongly correlates with portrait beauty ($\rho=0.12$), as well as the \textit{Overall\_Sharpness} metric.   As hypothesized, lighting patterns are also fundamental for a good portrait. This is shown by the positive $\rho$ of the face/background lighting contrast feature. Moreover, our analysis shows that there is a relation between image beauty and illumination patterns ( e.g. Clusters 3 has positive $\rho$, while Cluster 4 has negative  $\rho$). Overall, our new lighting features show higher relation with beauty than basic brightness features ($\rho=0.054$ for the \textit{Average\_V} features), confirming the need of more complex lighting features for portrait aesthetic evaluation. 
Similarly, contrast in colors and in gray levels (\textit{GLCM\_Contrast} and \textit{Contrast\_Michelson}) also show positive correlation with aesthetic scores.

Moreover, negative $\rho$ values for \textit{Noise}  and positive correlation with \textit{GLCM\_Energy} make us conclude that visually appealing portraits should have a homogeneous, smooth composition without disturbing distortions. We can also see that the amount of \textit{Median\_Filtering} is negatively correlated with beauty, showing that too intensive post-processing results in a decrease of the portrait appeal. Surprisingly, \textit{Exposure\_Quality} is negatively correlated with beauty, suggesting that playing with over/under exposure results in more appealing pictures. Moreover, negative $\rho$ for some Color Names indicates that beautiful portraits tend to have little regions colored with non-skin colors such as green, purple, magenta. We can also notice the good outcome of our attempt of modeling fuzzy properties, given that properties such as \textit{Originality} and \textit{Uniqueness} positively correlate with beauty.

It was very interesting to notice how physical/demographic properties such as gender, eye color, glasses, age, and race show very low correlation with image beauty, suggesting that any subject, no matter his/her traits, can be part of a stunning picture, if the photographer is able to grasp the subject's essence.

By correlating gender properties with other visual features, we could find some side curious insights about portraiture. For example, female pictures tend to be more memorable, as well as  brighter and post-processed, while male tend to be represented with darker colors, and smile less than females.
%

%% file: sec/6_results.tex
\section{Predicting Portrait Beauty}\label{results}
In order to test the effectiveness of our proposed features, and verify the findings of our analysis (see Sec. \ref{analysis}), we perform 2 different classification experiments. First, we perform a small-scale experiment on the dataset provided in \cite{khan2012evaluating}, showing the performances of our method and comparing them with the face-specific framework proposed by \cite{khan2012evaluating}. Then, we design a large-scale classification framework, by looking at the ability  of our features to discriminate between beautiful/non-beautiful pictures, using the large-scale dataset we built in Sec. \ref{datasets}. We compare the classification performances of a framework based on our different groups of features with the one of a generic aesthetic classifier, i.e. based on traditional compositional features and trained on images with diverse subjects.

 \subsection{Small-Scale Experiment}
The work that more closely relates to ours is the portrait aesthetic framework from Khan et al. \cite{khan2012evaluating}: they design face-specific features and computes their effectiveness on a publicly available small-scale dataset of 150 pictures. 
\input{tables/result_small}
In order to test the performances of our approach, we compute the visual features in Sec. \ref{visual_features} on the dataset from Khan et al. \cite{khan2012evaluating} and we prove their effectiveness by using the same experimental setup, i.e. binarization of scores based on median, 10-folds cross validation on an SVM classifier in Weka, and average accuracy as evaluation metric. For fair comparison, we first evaluate the classification performances on our portrait-specific features only (see Sec. \ref{facefeatures}), reporting results with and without feature selection in Table \ref{tab:small_experiments}. Our group of portrait features alone outperforms the system in \cite{khan2012evaluating}. Moreover, when we use all the features proposed in this work for the same classification task, we reach even higher classification accuracy, observing a substantial improvement of the performances compared to our baseline (and similar works such as the one from Li et al. \cite{li2010aesthetic}).
\subsection{Large-Scale Aesthetic Categorization}
We now test the proposed approach for aesthetic classification on a large-scale, using the dataset of Sec. \ref{datasets}.
To classify the images as ``Beautiful" and ``Non-beautiful", we use the binaries the average AVA 
scores it by labeling as positive any image with a score
greater than the mean user score ($5.55$). Similar to \cite{murray2012ava}, we learn a SVM classifier using the publicly available libSVM
package. For this, the dataset is randomly divided into 5 partitions, as in Sec. \ref{analysis}, and a SVM
classifier is learned per partitions. We use RBF kernel where the $\gamma$ parameter is set
to $1/n$ where $n$ is the number of features. The cost parameter $C$ is obtained
using 10-fold cross-validation. All features are standardized to be zero mean
and unit variance. 

Fig \ref{fig:analysis} (c)  shows the average classification accuracy on the test set for each group of features. As we can see, our framework benefits from the combination of diverse features, since the best performance is given by all features combined with early fusion, ($64.24\% \pm 1.76$) . Moreover, as expected by our analysis, we confirm that the classifier based on our rich portrait features outperforms the  classifiers based on the other groups of features, suggesting that detailed information of face properties and landmarks is more discriminative for portrait classification than traditional compositional features.

Results reported in \cite{murray2012ava, interesting} proved that a classifier trained on non-specific images performs better than a portrait-specific framework.  To prove the importance of building a portrait-specific framework, we compare our results with a baseline classifier built with traditional compositional features only (as in Sec. \ref{sec:comp}), and trained on the dataset used in \cite{pere}, namely  a database of  images belonging to 7 different categories, including ``Portraiture",``Flower", etc. and annotated with the corresponding aesthetic scores from DPchallenge.com (same source as our dataset, same score range). Unlike the findings in \cite{murray2012ava, interesting}, we confirm the hypothesis that portraits need a separate computational framework for aesthetic assessment, showing that all the classifiers based on our proposed features perform better than this baseline (with all features, the improvement is more than 16\%).

As in ~\cite{murray2012ava}, we also performed SVM classification by introducing a $\delta$
parameter to discard ambiguous images from the training set (keeping all the
images in the test set). The $\delta$ parameter was ranged from $0.1$ to $1.0$,
but unlike~\cite{murray2012ava} we did not experience any increase in the classification
accuracy. However, the performance with the $\delta = 0.5$ is similar to when
$\delta = 0.0$, implying that the ambiguous images do not help for the task of
classification and can be discarded to speed up the learning time.  

%% file: tables/result_small.tex
\begin{table}
\small
\centering
\resizebox{0.6\linewidth}{!}{
	\begin{tabular}{|l|c|c|}
\hline
\textbf{Feature} & \textbf{Dim} & \textbf{Accuracy} \\ \hline
\textit{Baseline \cite{khan2012evaluating}} &7& 61.10\% \\  
\textit{Face Features} &44& 62,88\%\\  
\textit{Face Features (sel)} &11& 68,94\%\\ 
\textit{Non-Face Features} &276& 65,15\% \\ 
\textit{Non Face Features (sel)} &9& 68,18\%\\ 
\textit{All Features)} &320& 66,67\%\\ 
\textit{All Features (sel)} &12& 75,76\%\\ \hline
\end{tabular}}
 \caption{Classification Accuracy on the dataset from \cite{khan2012evaluating}}
 \label{tab:small_experiments}
\vspace{-1cm}
 \end{table}

%% file: sec/7_conclusions.tex
\section{Conclusions}\label{conclusions}
In this paper, we presented a complete framework for large-scale portrait aesthetic assessment based on visual features. We procured a dataset of digital portraits annotated with aesthetic scores and other information regarding traits/demographics of the subjects in the portraits. We designed a set of discriminative visual features based on portrait photography literature. We analyzed the importance of each feature for portrait beauty, showing that rich facial features play a significant role in guiding the portrait aesthetics, and that the perceived portrait beauty is largely independent of the demographic characteristics of the subject. Finally, we built a classifier that is able to successfully distinguish between beautiful and non-beautiful portraits.

In our future work, we plan to broaden our framework by extending our database to include portrait images 'in the wild', exploring portrait aesthetics with a more challenging context.

%% file: FG2015_archive.bbl
\begin{thebibliography}{10}

\bibitem{facesinstagram}
Bakhshi, S., Shamma, D., Gilbert, E.:
\newblock Faces engage us: Photos with faces attract more likes and comments on
  instagram.
\newblock  (In: CHI 2014)

\bibitem{child2008studio}
Child, J.:
\newblock Studio photography: essential skills.
\newblock CRC Press (2008)

\bibitem{hurter2007portrait}
Hurter, B.:
\newblock Portrait Photographer's Handbook.
\newblock Amherst Media, Inc (2007)

\bibitem{weiser1999phototherapy}
Weiser, J.:
\newblock Phototherapy techniques: Exploring the secrets of personal snapshots
  and family albums.
\newblock PhotoTherapy Centre (1999)

\bibitem{datta}
Datta, R., Joshi, D., Li, J., Wang, J.:
\newblock Studying aesthetics in photographic images using a computational
  approach.
\newblock ECCV (2006)  288--301

\bibitem{ke2006design}
Ke, Y., Tang, X., Jing, F.:
\newblock The design of high-level features for photo quality assessment.
\newblock In: CVPR. Volume~1., IEEE (2006)  419--426

\bibitem{khan2012evaluating}
Khan, S.S., Vogel, D.:
\newblock Evaluating visual aesthetics in photographic portraiture.
\newblock In: CaE, Eurographics Association (2012)  55--62

\bibitem{li2010aesthetic}
Li, C., Gallagher, A., Loui, A.C., Chen, T.:
\newblock Aesthetic quality assessment of consumer photos with faces.
\newblock In: ICIP, 2010, IEEE (2010)  3221--3224

\bibitem{murray2012ava}
Murray, N., Marchesotti, L., Perronnin, F.:
\newblock Ava: A large-scale database for aesthetic visual analysis.
\newblock In: CVPR, 2012 IEEE Conference on, IEEE (2012)  2408--2415

\bibitem{isola2011}
Isola, P., Xiao, J., Torralba, A., Oliva, A.:
\newblock What makes an image memorable?
\newblock In: CVPR, IEEE (2011)  145--152

\bibitem{emotions}
Machajdik, J., Hanbury, A.:
\newblock Affective image classification using features inspired by psychology
  and art theory.
\newblock In: MM, ACM (2010)  83--92

\bibitem{joshi2011aesthetics}
Joshi, D., Datta, R., Fedorovskaya, E., Luong, Q.T., Wang, J.Z., Li, J., Luo,
  J.:
\newblock Aesthetics and emotions in images.
\newblock Signal Processing Magazine, IEEE \textbf{28} (2011)  94--115

\bibitem{redi2012interestingness}
Redi, M., Merialdo, B.:
\newblock Where is the interestingness?: retrieving appealing videoscenes by
  learning flickr-based graded judgments.
\newblock In: MM, ACM (2012)  1363--1364

\bibitem{interesting}
Dhar, S., Ordonez, V., Berg, T.:
\newblock High level describable attributes for predicting aesthetics and
  interestingness.
\newblock In: CVPR, IEEE (2011)  1657--1664

\bibitem{gygli2013interestingness}
Gygli, M., Grabner, H., Riemenschneider, H., Nater, F., Van~Gool, L.:
\newblock The interestingness of images.
\newblock (2013)

\bibitem{zerr2012privacy}
Zerr, S., Siersdorfer, S., Hare, J., Demidova, E.:
\newblock Privacy-aware image classification and search.
\newblock In: ACM SIGIR conference on Research and development in information
  retrieval, ACM (2012)  35--44

\bibitem{nishiyama2011aesthetic}
Nishiyama, M., Okabe, T., Sato, I., Sato, Y.:
\newblock Aesthetic quality classification of photographs based on color
  harmony.
\newblock In: CVPR, IEEE (2011)  33--40

\bibitem{marchesotti2011assessing}
Marchesotti, L., Perronnin, F., Larlus, D., Csurka, G.:
\newblock Assessing the aesthetic quality of photographs using generic image
  descriptors.
\newblock In: ICCV, IEEE (2011)  1784--1791

\bibitem{wu2011learning}
Wu, O., Hu, W., Gao, J.:
\newblock Learning to predict the perceived visual quality of photos.
\newblock In: ICCV, IEEE (2011)  225--232

\bibitem{bhattacharya2010framework}
Bhattacharya, S., Sukthankar, R., Shah, M.:
\newblock A framework for photo-quality assessment and enhancement based on
  visual aesthetics.
\newblock In: MM, ACM (2010)  271--280

\bibitem{luo2011content}
Luo, W., Wang, X., Tang, X.:
\newblock Content-based photo quality assessment.
\newblock In: ICCV), 2011, IEEE (2011)  2206--2213

\bibitem{xuefeature}
Xue, S.F., Tang, H., Tretter, D., Lin, Q., Allebach, J.:
\newblock
\newblock (Feature design for aesthetic inference on photos with faces)

\bibitem{face++}
Inc., M.:
\newblock Face++ research toolkit.
\newblock www.faceplusplus.com (2013)

\bibitem{chakrabarti2010analyzing}
Chakrabarti, A., Zickler, T., Freeman, W.T.:
\newblock Analyzing spatially-varying blur.
\newblock In: CVPR, 2010 IEEE Conference on, IEEE (2010)  2512--2519

\bibitem{Michel}
Michelson, A.:
\newblock Studies in optics.
\newblock Dover Pubns (1995)

\bibitem{order}
Rigau, J., Feixas, M., Sbert, M.:
\newblock Conceptualizing birkhoff's aesthetic measure using shannon entropy
  and kolmogorov complexity.
\newblock Computational Aesthetics in Graphics, Visualization, and Imaging
  (2007)

\bibitem{li2010object}
Li, L.J., Su, H., Xing, E.P., Li, F.F.:
\newblock Object bank: A high-level image representation for scene
  classification \& semantic feature sparsification.
\newblock In: NIPS. Volume~2. (2010) ~5

\bibitem{buades2005non}
Buades, A., Coll, B., Morel, J.M.:
\newblock A non-local algorithm for image denoising.
\newblock In: CVPR 2005. IEEE Computer Society Conference on. Volume~2., IEEE
  (2005)  60--65

\bibitem{wang2002no}
Wang, Z., Sheikh, H.R., Bovik, A.C.:
\newblock No-reference perceptual quality assessment of jpeg compressed images.
\newblock In: Image Processing. 2002. Proceedings. 2002 International
  Conference on. Volume~1., IEEE (2002)  I--477

\bibitem{borth2013large}
Borth, D., Ji, R., Chen, T., Breuel, T., Chang, S.F.:
\newblock Large-scale visual sentiment ontology and detectors using adjective
  noun pairs.
\newblock In: ACM international conference on Multimedia, ACM (2013)  223--232

\bibitem{dan2011geneva}
Dan-Glauser, E.S., Scherer, K.R.:
\newblock The geneva affective picture database (gaped): a new 730-picture
  database focusing on valence and normative significance.
\newblock Behavior research methods \textbf{43} (2011)  468--477

\bibitem{pere}
Obrador, P., Saad, M., Suryanarayan, P., Oliver, N.:
\newblock Towards category-based aesthetic models of photographs.
\newblock Advances in Multimedia Modeling (2012)  63--76

\bibitem{gross2003image}
Gross, R., Brajovic, V.:
\newblock An image preprocessing algorithm for illumination invariant face
  recognition.
\newblock In: Audio-and Video-Based Biometric Person Authentication, Springer
  (2003)  10--18

\bibitem{ng2001practical}
Ng~Kuang~Chern, N., Neow, P.A., Ang, V.:
\newblock Practical issues in pixel-based autofocusing for machine vision.
\newblock In: ICRA. IEEE International Conference on. Volume~3., IEEE (2001)
  2791--2796

\bibitem{dalal2005histograms}
Dalal, N., Triggs, B.:
\newblock Histograms of oriented gradients for human detection.
\newblock In: CVPR 2005. IEEE Computer Society Conference on. Volume~1., IEEE
  (2005)  886--893

\bibitem{hou2007saliency}
Hou, X., Zhang, L.:
\newblock Saliency detection: A spectral residual approach.
\newblock In: CVPR, IEEE (2007)  1--8

\bibitem{pevny2010steganalysis}
Pevny, T., Bas, P., Fridrich, J.:
\newblock Steganalysis by subtractive pixel adjacency matrix.
\newblock information Forensics and Security, IEEE Transactions on \textbf{5}
  (2010)  215--224

\bibitem{casia}
:
\newblock
\newblock (Credits for the use of the casia image tempering detection
  evaluation database (caisa tide) v1.0 are given to the national laboratory of
  pattern recognition, institute of automation, chinese academy of science,
  corel image database and the photographers. http://forensics.idealtest.org)

\bibitem{yuan2011blind}
Yuan, H.D.:
\newblock Blind forensics of median filtering in digital images.
\newblock Information Forensics and Security, IEEE Transactions on \textbf{6}
  (2011)  1335--1345

\bibitem{sm}
Redi, M., Merialdo, B.:
\newblock Saliency moments for image categorization.
\newblock In: ICMR. (2011)

\bibitem{tibshirani1996regression}
Tibshirani, R.:
\newblock Regression shrinkage and selection via the lasso.
\newblock Journal of the Royal Statistical Society. Series B (Methodological)
  (1996)  267--288

\end{thebibliography}
